\pdfoutput=1

\documentclass[11pt]{article}

\usepackage[preprint]{acl}

\usepackage{times}
\usepackage{latexsym}
\usepackage{algorithm}
\usepackage{algpseudocode}
\usepackage{amsmath}
\usepackage{multirow}
\usepackage{array}
\usepackage{multirow}
\usepackage{graphicx}
\usepackage{colortbl}
\usepackage{amssymb}
\usepackage{pifont} 
\usepackage{booktabs}
\usepackage{subcaption}  
\usepackage{caption}    

\usepackage[T1]{fontenc}

\usepackage[utf8]{inputenc}

\usepackage{microtype}

\usepackage{inconsolata}

\usepackage{graphicx}

\title{\texttt{1bit}-Merging: Dynamic Quantized Merging for Large Language Models}

\author{Shuqi Liu$^{1,2}$, Yuxuan Yao$^{1}$, Bowei He$^{1}$, Zehua Liu$^2$, Xiongwei Han$^2$, Mingxuan Yuan$^2$\\
\textbf{Han Wu$^{2,}$$^\dagger$, Linqi Song$^{1,}$$^\dagger$}\\
$^{1}$ Department of Computer Science, City University of Hong Kong\\
$^{2}$ Huawei Noah's Ark Lab\\
\texttt{shuqiliu4-c@my.cityu.edu.hk}\\
\texttt{wu.han1@huawei.com}\\
\texttt{linqi.song@cityu.edu.hk}
}

\begin{document}
\maketitle
\begin{abstract}
Recent advances in Large Language Models (LLMs) have led to specialized models excelling in specific domains, creating a need for efficient model merging techniques. While traditional merging approaches combine parameters into a single static model, they often compromise task-specific performance. However, task-specific routing methods maintain accuracy but introduce substantial storage overhead. We present \texttt{1bit}-Merging, a novel framework that integrates task-specific routing with 1-bit quantized task vectors to balance performance and storage efficiency. Our approach further leverages the observation that different task-specific models store knowledge in distinct layers, such as MLP and attention layers, enabling targeted compression strategies. Through extensive experiments with LLaMA2 and Mistral model families across general knowledge, mathematical reasoning, and code generation tasks, 
we demonstrate that \texttt{1bit}-Merging not only outperforms traditional model merging methods but also achieves better storage efficiency than task-specific routing approaches, maintaining \textbf{94.53\%} of their performance while reducing storage requirements to \textbf{55.02\%} on merging Chat, Math and Code fine-tuned models on Mistral 7B models. 
\end{abstract}

\begin{figure}[t]
    \centering
    \includegraphics[width=0.9\linewidth]{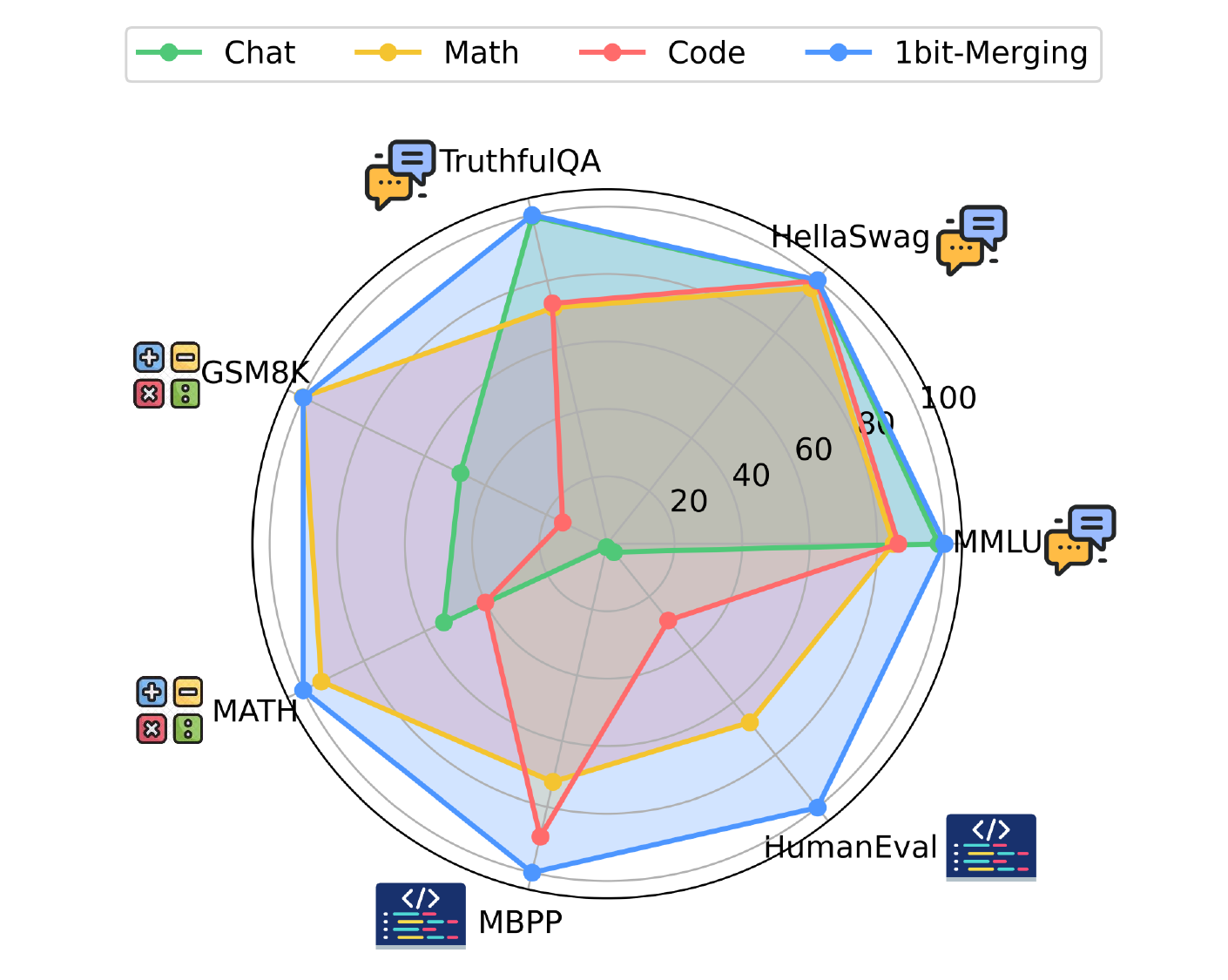}
    \caption{While individually fine-tuned models excel only in their specialized domains, our \texttt{1bit}-Merging achieves superior performance across all domains.}
    \label{fig:radar_combined}
    \vspace{-0.5cm}
\end{figure}

{
\let\thefootnote\relax\footnotetext{
$^\dagger$Corresponding author.}
}
\section{Introduction}
Large language models (LLMs) have achieved remarkable progress, demonstrating favourable performance on a wide range of tasks \cite{touvron2023llama, zhao2023survey}. As a variety of fine-tuned (FT) models for specific domains have emerged, there is a growing need to combine their specialized capabilities into a single model \citep{merge_survey, mergeKit}. While multi-task learning offers one solution \cite{sanh2022multitask, fifty2021efficiently}, it requires extensive computational resources and simultaneous access to all task-specific datasets. Recent advances in parameter-space model merging \cite{model_soup, task_arithmetic, TIES_merging, dare} provide an efficient alternative, which combines pre-trained (PT) and FT models through parameter manipulation without additional training overhead, resulting in a unified model with comprehensive capabilities. 

Within the framework of model merging, task vectors \cite{task_arithmetic, TIES_merging, dare} are pivotal for encoding task-specific knowledge. Defined as parametric deltas between PT and FT model weights, these vectors theoretically facilitate cross-model capability integration via linear arithmetic operations. Nonetheless, traditional merging often sacrifices task-specific performance, resulting in a noticeable gap compared to individual expert models. For instance, merging the \textit{math-specific}, and \textit{code-specific} models may result in a model that performs better in math than the \textit{code-specific} model, but worse in code tasks.
To alleviate such challenges, merging with task-specific routing \citep{smear, twin-merging} dynamically prioritizes relevant task vectors based on input data, effectively maintaining accuracy by isolating task-specific parameters. However, this routing-based merging strategy introduces substantial storage overhead, as it necessitates the preservation of all task vectors. SVD-based and quantization-based approaches are proposed to resolve the problem \cite{wang2025dobi, liu2024bitdeltafinetuneworthbit}, but require complex high-dimensional computations or utilize the same quantization method for all components, contradicting previous observations \cite{xu2025scalablemodelmergingprogressive,liu2025sensmergingsensitivityguidedparameterbalancing} that different layers possess varying importance.  
Building upon earlier studies that highlight the varying sensitivity across layers \citep{liu2025sensmergingsensitivityguidedparameterbalancing}, we further investigate and identify differing levels of importance among functional modules, such as MLP and Attention, which can be leveraged for model merging. Notably, our empirical findings reveal that task-specific models encode knowledge in distinct modules. For example, chat-oriented models predominantly store knowledge within attention layers, enabling the compression of MLP layers. In contrast, math- and code-related models primarily encode knowledge in MLP layers, allowing for the compression of attention layers.

In this work, we introduce \texttt{1bit}-Merging, a novel dynamic merging framework that integrates task-specific routing with adaptive 1-bit quantized task vectors. 
Recognizing the substantial redundancy inherent within task vectors, we implement 1-bit compression, which significantly reduces storage requirements without notably compromising model capacities.
Building upon the compressed task vectors, our method employs task-specific routing to establish a task-specific base model. This base model serves as the foundation for integrating the remaining compressed task vectors, ensuring that each task leverages the most relevant and efficient parameters.
\texttt{1bit}-Merging thus offers a balanced solution that maintains the performance advantages of task-specific routing while addressing the storage inefficiencies in existing methods.


Empirically, we conducted extensive experiments by comparing \texttt{1bit}-Merging with existing model merging approaches. We merged three widely used fine-tuned models focused on general knowledge (Chat), mathematical reasoning (Math), and code generation (Code) from the LLaMA2 and Mistral families. Our results show that \texttt{1bit}-Merging outperforms traditional model merging methods while offering better storage efficiency than task-specific routing approaches. The task vector compression method preserves and often enhances the capabilities of FT models. By combining these compressed task vectors with our dynamic routing strategy, \texttt{1bit}-Merging achieves superior performance. As shown in Figure \ref{fig:radar_combined}, it surpasses individually fine-tuned models across all domains.


To sum up, our contributions include:
(1) We discover that different task-specific models store knowledge in distinct layers, enabling targeted compression strategies based on specific compression position.
(2) We propose a novel dynamic merging framework that integrates task-specific routing with 1-bit quantized task vectors.
(3) Through comprehensive evaluations, 
we demonstrate that \texttt{1bit}-Merging not only outperforms conventional merging methods but also achieves superior storage efficiency compared to task-specific routing. It retains \textbf{94.53\%} of the performance of full-precision models while reducing storage requirements to \textbf{55.02\%}, when merging on Mistral 7B models.

\section{Related Work}
Model merging, a paradigm to enhance model generalization and robustness by synthesizing heterogeneous neural networks, is broadly categorized into static and dynamic approaches. 

\paragraph{Static Model Merging}
Static model merging primarily explores general strategies for combining models, such as Average Merging \citep{model_soup}, Task Arithmetic \citep{task_arithmetic}, and TIES-Merging \citep{TIES_merging}. Notably, \citet{model_soup} first demonstrated that even a straightforward weight averaging of base models can improve both the performance and robustness of downstream tasks.
Building on this, Task Arithmetic \citep{task_arithmetic} refines the merging process by introducing task vectors, proposing that simple arithmetic operations on these vectors can effectively modify models and yield a better merged model. Expanding upon the concept of task vectors, methods like DARE \citep{dare} and TIES-Merging \citep{TIES_merging} adopt pruning-then-scaling techniques to merge task vectors, based on the premise that not all parameters equally contribute to the final performance. 
However, static merging of models from different domains often sacrifices task-specific performance to strike a balance between generalization capacity and task-specific effectiveness. To this end, we explore the dynamic model merging in this work.

\paragraph{Dynamic Model Merging}

Additionally, another line of studies focuses on routing-based model merging.
For instance, SMEAR \citep{smear} propose a routing-based merging paradigm where parameter fusion is implemented through weighted averaging guided by router input distributions across expert modules, maintaining computational efficiency comparable to singular expert operations.
Extending this paradigm, Twin-Merging \citep{twin-merging} develop an adaptive knowledge integration framework that dynamically reconciles task-shared and task-specific representations via routing mechanisms during inference.
\citep{weight-ensemble} advance this domain through a Transformer-based dynamic composition architecture, with empirical analysis revealing disproportionate parameter modification magnitudes between linear and nonlinear layers during fine-tuning - a critical factor affecting integration efficacy. Notwithstanding their methodological advancements, extant routing-based merging frameworks incur significant storage demands for task vector retention, which motivates the development of our computationally efficient model merging framework in this work.

\section{Method}

\subsection{Preliminaries}

\paragraph{Task Vectors}


Given a set of $ K $ fine-tuned models $ \{\theta^{t_1}_{\text{SFT}}, \theta^{t_2}_{\text{SFT}}, \ldots, \theta^{t_K}_{\text{SFT}}\} $ derived from a shared pre-trained backbone $ \theta_{\text{PRE}} $, each task vector is defined as the parameter difference between the fine-tuned and pre-trained models:
\begin{equation}
    \Delta \theta^{t_k} = \theta^{t_k}_{\text{SFT}} - \theta_{\text{PRE}}.
\end{equation}
Traditional merging methods combine all task vectors to produce a single, static merged model:
\begin{equation}
    \theta_{\text{merged}}^{\text{traditional}} = \theta_{\text{PRE}} + \sum_{k=1}^{K} \delta_{t_k}.
\end{equation}
In contrast, routing-based merging dynamically selects a task-specific vector based on the input $ \mathbf{x} $ using a router function:
\begin{equation}
    \theta_{\text{merged}}^{\text{routing}} = \theta_{\text{PRE}} + \delta_{t_i}, \quad i = \mathop{\mathrm{argmax}}\limits_{k}~\mathrm{Router}(\mathbf{x}),
\end{equation}
\noindent where $ t_i $ denotes the task vector selected by the well-trained router.


\begin{figure}[htbp]
    \centering
    \includegraphics[width=0.85\linewidth]{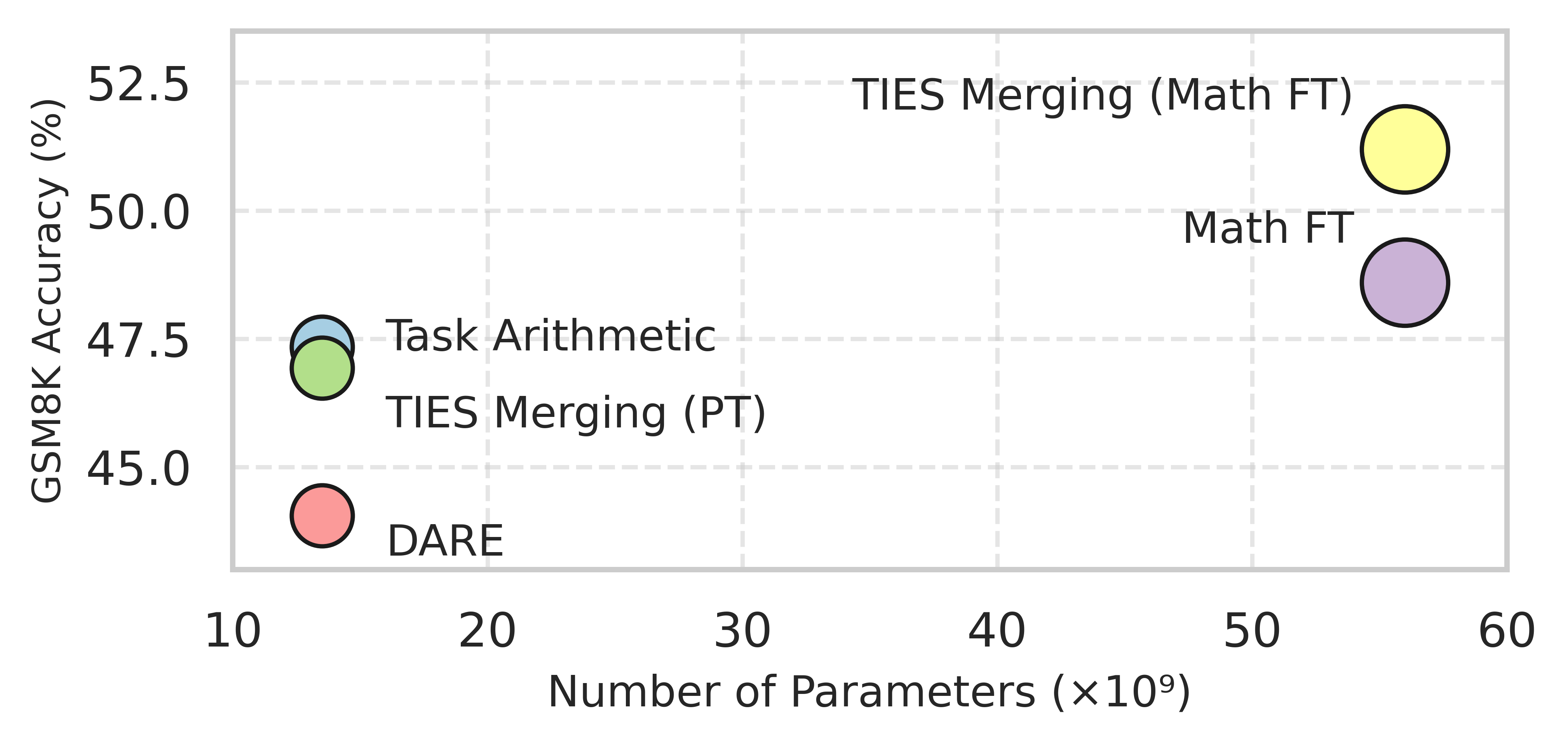}
    \vspace{-3mm}
    \caption{Performance comparison of different strategies on GSM8K on LLaMA2-7B series. TIES-Merging achieves superior performance when choosing Math FT model as base model.}
    \label{fig:gsm8k_acc}
    \vspace{-5mm}
\end{figure}

\paragraph{Impact of Base Model}
Leveraging the advantages of task-specific routing, we further investigate how the choice of base model affects model merging performance. Specifically, we compare the use of a pre-trained (PT) backbone $ \theta_{\text{PRE}} $ versus a fine-tuned (FT) model for mathematics $ \theta_{\text{Math}} = \theta_{\text{PRE}} + \delta_{t_{\text{Math}}} $, as the base model in mathematical tasks. As shown in Figure~\ref{fig:gsm8k_acc}, substituting the PT backbone with a math-specialized FT model (Math-FT) yields substantial performance improvements on the GSM8K dataset~\cite{cobbe2021training} when used in conjunction with the TIES-Merging approach. This result highlights the importance of selecting an appropriate base model for effective model merging.

Despite outperforming individually fine-tuned models, the use of a task-specific base model introduces increased storage overhead, as multiple task vectors requires to be loaded concurrently.




\subsection{Unveiling the Varying Module Importance} \label{unveiling}

Inspired by prior work on module-level and layer-level importance in LLMs \cite{meng2022locating, liu2025sensmergingsensitivityguidedparameterbalancing}, we investigate how different expert models respond to 1-bit quantization across various architectural components, including the Attention module, MLP module, and all Linear layers.

As shown in Table \ref{tab:ablation}, for Chat model, performance remains largely preserved after quantizing the MLP module, suggesting that instruction-following capabilities are primarily encoded in the Attention module. In contrast, for Math model, quantizing the Attention module has minimal impact, while quantizing MLP leads to significant degradation—indicating that complex reasoning capabilities are predominantly stored in MLP subnetworks. Similar phenomenon can be observed in Table \ref{tab:position_mistral} and Table \ref{tab:position_llama_13} for Mistral-7B and LLaMA2-13B models correspondingly.
These findings highlight that knowledge localization varies across modules in a task-dependent manner, and that applying a task-agnostic compression strategy—such as uniform quantization—can severely degrade performance on complex reasoning tasks.


\begin{table*}[t]
\renewcommand{\arraystretch}{0.95}
\resizebox{\textwidth}{!}{%
\begin{tabular}{lcccclcclccc}
\toprule
 &
   &
  \multicolumn{3}{c}{General Knowledge} &
   &
  \multicolumn{2}{c}{Mathmetical Reasoning} &
   &
  \multicolumn{2}{c}{Code Generation} &
  \multicolumn{1}{c}{} \\ \cline{3-5} \cline{7-8} \cline{10-11}
\multirow{-2}{*}{Models} &
  \multirow{-2}{*}{Position} &
  \multicolumn{1}{c}{MMLU} &
  \multicolumn{1}{c}{HellaSwag} &
  \multicolumn{1}{c}{TruthfulQA} &
   &
  \multicolumn{1}{c}{GSM8K} &
  \multicolumn{1}{c}{MATH} &
   &
  \multicolumn{1}{c}{MBPP} &
  \multicolumn{1}{c}{HumanEval} &
  \multicolumn{1}{c}{\multirow{-2}{*}{Average}} \\
  \hline
 &
  \cellcolor[HTML]{EFEFEF}/ &
  \cellcolor[HTML]{EFEFEF}\underline{46.38} &
  \cellcolor[HTML]{EFEFEF}57.79 &
  \cellcolor[HTML]{EFEFEF}\textbf{45.17} &
  \cellcolor[HTML]{EFEFEF} &
  \cellcolor[HTML]{EFEFEF}23.43 &
  \cellcolor[HTML]{EFEFEF}4.86 &
  \cellcolor[HTML]{EFEFEF} &
  \cellcolor[HTML]{EFEFEF}0.3 &
  \cellcolor[HTML]{EFEFEF}0.6 &
  \cellcolor[HTML]{EFEFEF}\textbf{25.50} \\
 &
  Attention &
   42.12&
   55.30&
   41.13&
   &
   22.74&
   5.08&
   &
   0.0&
   0.0&
   23.70\\
 &
  MLP &
   \cellcolor[HTML]{CBCEFB}\textbf{46.40}&
   \cellcolor[HTML]{CBCEFB}\underline{58.23}&
   \cellcolor[HTML]{CBCEFB}\underline{43.70}&
   &
   21.15&
   4.94&
   &
   0.0&
   0.0&
   \underline{25.30}\\
\multirow{-4}{*}{Chat} &
  Linear &
   45.61&
   \textbf{58.38}&
   42.96&
   &
   20.70&
   4.76&
   &
   0.0&
   0.0&
   24.63\\
   \hline
 &
  \cellcolor[HTML]{EFEFEF}/ &
  \multicolumn{1}{c}{\cellcolor[HTML]{EFEFEF}40.05} &
  \multicolumn{1}{c}{\cellcolor[HTML]{EFEFEF}56.30} &
  \multicolumn{1}{c}{\cellcolor[HTML]{EFEFEF}32.56} &
  \cellcolor[HTML]{EFEFEF} &
  \multicolumn{1}{c}{\cellcolor[HTML]{EFEFEF}\textbf{48.60}} &
  \multicolumn{1}{c}{\cellcolor[HTML]{EFEFEF}\textbf{8.50}} &
  \cellcolor[HTML]{EFEFEF} &
  \multicolumn{1}{c}{\cellcolor[HTML]{EFEFEF}21.8} &
  \multicolumn{1}{c}{\cellcolor[HTML]{EFEFEF}12.8} &
  \multicolumn{1}{c}{\cellcolor[HTML]{EFEFEF}31.52} \\
 &
  Attention &
  \multicolumn{1}{c}{41.03} &
  \multicolumn{1}{c}{56.68} &
  \multicolumn{1}{c}{34.52} &
   &
  \cellcolor[HTML]{CBCEFB}{\underline{47.46}} &
  \cellcolor[HTML]{CBCEFB}{\underline{8.42}} &
   &
  \multicolumn{1}{c}{23.1} &
  \multicolumn{1}{c}{\textbf{14.0}} &
  \multicolumn{1}{c}{\textbf{32.17}} \\
 &
  MLP &
  \multicolumn{1}{c}{41.85} &
  \multicolumn{1}{c}{57.68} &
  \multicolumn{1}{c}{33.17} &
   &
  \multicolumn{1}{c}{44.88} &
  \multicolumn{1}{c}{7.06} &
   &
  \multicolumn{1}{c}{25.3} &
  \multicolumn{1}{c}{14.0} &
  \multicolumn{1}{c}{{31.99}} \\
\multirow{-4}{*}{Math} &
  Linear &
  \multicolumn{1}{c}{42.76} &
  \multicolumn{1}{c}{57.96} &
  \multicolumn{1}{c}{34.52} &
   &
  \multicolumn{1}{c}{42.29} &
  \multicolumn{1}{c}{6.84} &
   &
  \multicolumn{1}{c}{24.3} &
  \multicolumn{1}{c}{15.2} &
  \multicolumn{1}{c}{31.98} \\
  \hline
 &
  \cellcolor[HTML]{EFEFEF}/ &
  \cellcolor[HTML]{EFEFEF}40.76 &
  \cellcolor[HTML]{EFEFEF}57.87&
  \cellcolor[HTML]{EFEFEF}33.17&
  \cellcolor[HTML]{EFEFEF} &
  \cellcolor[HTML]{EFEFEF}7.13&
  \cellcolor[HTML]{EFEFEF}3.62&
  \cellcolor[HTML]{EFEFEF} &
  \cellcolor[HTML]{EFEFEF}\underline{26.8}&
  \cellcolor[HTML]{EFEFEF}5.5&
  \cellcolor[HTML]{EFEFEF}{24.98}\\
 &
  Attention &
  \multicolumn{1}{c}{41.36} &
  \multicolumn{1}{c}{57.81} &
  \multicolumn{1}{c}{32.80} &
   &
  \multicolumn{1}{c}{8.19} &
  \multicolumn{1}{c}{3.36} &
   &
  \cellcolor[HTML]{CBCEFB}{\textbf{27.3}} &
  \cellcolor[HTML]{CBCEFB}{\underline{11.0}} &
  \multicolumn{1}{c}{\textbf{25.97}} \\
 &
  MLP &
  \multicolumn{1}{c}{41.08} &
  \multicolumn{1}{c}{57.80} &
  \multicolumn{1}{c}{33.29} &
   &
  \multicolumn{1}{c}{5.23} &
  \multicolumn{1}{c}{3.16} &
   &
  \multicolumn{1}{c}{14.0} &
  \multicolumn{1}{c}{2.4} &
  \multicolumn{1}{c}{22.42} \\
\multirow{-4}{*}{Code} &
  Linear &
   41.50&
   57.60&
   33.54&
   &
   7.51&
   3.06&
   &
   0.0&
   0.0& 20.46\\
   \bottomrule
\end{tabular}%
}
\caption{Impact of quantizing different module types on the performance of compressed models across Math, Code, and Chat expert variants derived from the LLaMA-2 7B backbone. Compression of attention modules yields the best results for Math and Code models, while MLP module compression performs optimally for the Chat model.}
\label{tab:ablation}
\vspace{-0.3cm}
\end{table*}

\subsection{\texttt{1bit}-Merging}
Based on the aforementioned analysis, we introduce \textbf{\texttt{1bit}-Merging} to balance performance with storage efficiency, which is a dynamic merging framework that combines task-specific routing with 1-bit quantized task vectors. According to the discussion in Section \ref{unveiling}, we only quantify specific modules, the workflow is outlined as follows.

\paragraph{1-bit Quantization of Task Vectors}
Recognizing the substantial redundancy within task vectors, we apply 1-bit weight quantization to the task vectors, converting the weight matrices in specific modules from FP32/16 precision to a 1-bit format. In this quantization process, each element of the task vector is set to either +1 or -1. To maintain performance despite the aggressive compression, we scale the binary weight matrices with scalar values in FP16 format. This scaling ensures that the quantized weights preserve the original weight \( L_2 \) norm, thereby maintaining model performance after the extremely low-bit compression \cite{liu2024bitdeltafinetuneworthbit}. The scaling factor \(\alpha\) is computed as:
\begin{equation}
\alpha = \frac{\|\mathbf{W}\|_1}{{m \cdot n}}, 
\end{equation}
\noindent where $m$ and $n$ are the dimensions of the weight matrix. Using this scaling factor, the transformed task vector $\tilde{\delta}_{t_k}$ is defined as:
\begin{equation}
    \tilde{\delta}_{t_k} = \alpha * \text{Sign}(\delta_{t_k})
\end{equation}


\paragraph{Dynamic Routing and Merging}
\label{para:dynamic_routing}

Our merging framework dynamically selects and integrates compressed task vectors through a task-aware routing mechanism. Given an input $\mathbf{x}$, a lightweight router—implemented as a three-layer linear network with Leaky ReLU activations and batch normalization—produces a softmax-normalized probability distribution over tasks:
\begin{equation}
    \mathbf{p} = \text{softmax}(f_{\theta}(\mathbf{x})),
\end{equation}
where $f_{\theta}$ denotes the router's neural architecture. The dominant task category is selected via:
\begin{equation}
    k^* = \mathrm{argmax}(\mathbf{p}),
\end{equation}
identifying the most relevant task vector $\tilde\delta_{t_{k^*}}$.
This vector is then added to the pre-trained backbone $\theta_{\text{PRE}}$ to construct a task-specific base model:
\begin{equation}
    \theta_{\text{base}} = \theta_{\text{PRE}} + \tilde\delta_{t_{k^*}}.
\end{equation}

Subsequently, we apply TIES-Merging~\cite{TIES_merging} to further integrate the remaining compressed task vectors into $\theta_{\text{base}}$, yielding a final merged model that adapts dynamically to the input while preserving overall performance through efficient parameter fusion:
\begin{equation}
    \mathbf{M}_{\text{merged}} = \text{TIES}\left(\theta_{\text{base}}, \{\delta_{t_k} \mid k \neq k^*\}_{k=1}^{K} \right).
\end{equation}
We use TIES-Merging to integrate task vectors, as it effectively resolves parameter conflicts via sign-based consensus, which aligns well with our use of 1-bit quantized task vectors.


\begin{table*}[!ht]
\renewcommand{\arraystretch}{0.95}
\resizebox{\textwidth}{!}{%
\begin{tabular}{lccclcclccc}
\toprule
\multirow{2}{*}{Method} &
  \multicolumn{3}{c}{General Knowledge} &
   &
  \multicolumn{2}{c}{Mathmetical Reasoning} &
   &
  \multicolumn{2}{c}{Code Generation} &
  \multirow{2}{*}{Average} \\ 
  \cline{2-4} \cline{6-7} \cline{9-10}
                & MMLU  & HellaSwag & TruthfulQA &  & GSM8K & MATH &  & MBPP & HumanEval &  \\
                \hline
                \rowcolor{gray!10} 
\multicolumn{11}{c}{\textit{Finetuned Models}} \\
Chat & \textbf{46.38} & 57.79 & \textbf{45.17} &  & 23.43 & 4.86 &  & 0.3  & 0.6 &  25.50\\
Math & 40.05 & 56.30 & 32.56 &  & \textbf{48.60} & \textbf{8.50} &  & 21.8 & \textbf{12.8} & \textbf{31.52} \\
Code & 40.76 & \textbf{57.87} & 33.17 &  & 7.13  & 3.62 &  & \textbf{26.8} & 5.5 & 24.98 \\
\hline
\rowcolor{gray!10} 
\multicolumn{11}{c}{\textit{Compressed Expert Models}} \\
\texttt{1bit}-Chat & \textbf{46.40} & \textbf{58.23} & \textbf{43.70} &  & 22.74 & 5.08 &  & 0.0  & 0.0 &  25.55\\
\texttt{1bit}-Math & 41.03 & 56.68 & 34.52 &  & \textbf{47.46} & \textbf{8.42} &  & 23.1 & \textbf{14.0} & \textbf{32.17} \\
\texttt{1bit}-Code & 41.36 & 57.81 & 32.80 &  & 8.19  & 3.36 &  &\textbf{27.3} & 11.0 & 25.97 \\
\hline
\rowcolor{gray!10} 
\multicolumn{11}{c}{\textit{Model Merging Methods}} \\
Task Arithmetic & 41.50 & 49.63 & 37.45 &  & 47.34 & 6.46 &  & 13.5 & 7.3 &  29.03\\
TIES-Merging    & 45.75 & 56.63 & 40.02 &  & 46.93 & 7.74 &  & 29.1 & 17.1 &  34.75\\
DARE & 46.81 & 57.57 & 38.19 &  & 44.05 & 6.98 &  & \textbf{31.6} & \textbf{18.9} & 34.87 \\
Twin-Merging          & \textbf{48.36} & \textbf{58.23} & 45.17 & & 46.43 & 8.14 & & 29.1 & 13.3 & 35.53 \\
\rowcolor{gray!20}
\texttt{1bit}-Merging    & 47.23 & 58.04 & \textbf{45.37} &  & \textbf{48.52} & \textbf{9.04} &  & 30.1 & \textbf{18.9} & \textbf{36.74} \\
\bottomrule
\end{tabular}%
}
\vspace{-0.1cm}
\caption{Performance evaluation of merged LLaMA2-7B Models (Chat, Math, Code) across 7 task-specific datasets.}
\vspace{-0.5cm}
\label{tab:llama2-7b}
\end{table*}

\section{Experiments}

In this section, we present a comprehensive evaluation of our \texttt{1bit}-Merging method from multiple perspectives. We begin by comparing our approach to expert models and traditional model merging techniques, aiming to assess its ability to preserve and effectively combine knowledge across different tasks. We then evaluate its performance against dynamic merging methods, focusing on adaptive and task-aware fusion scenarios. Finally, we examine the scalability of our method with respect to model architecture and parameter size, demonstrating its broader applicability across diverse models.

\subsection{Experimental Setup}

\paragraph{Baselines}
We consider both traditional and dynamic model merging methods as baselines. For traditional approaches, we include Task Arithmetic \citep{task_arithmetic}, which linearly combines task-specific models in the weight space; TIES-Merging \cite{TIES_merging}, which identifies consensus directions for parameter fusion; and DARE \cite{dare}, which randomly prunes parameters before merging to improve generalization. As a representative dynamic merging method, we adopt Twin-Merging \cite{twin-merging}, which introduces a trainable router to adaptively combine models based on input features. In particular, Twin Merging employs singular value decomposition (SVD) to compress the parameter space, enabling efficient and expressive model interpolation. We refer readers to Appendix~\ref{appen:implement_detail} for detailed implementation of our router, and hyperparameter settings for the merging methods.


\begin{table*}[]
\renewcommand{\arraystretch}{0.95}
\resizebox{\textwidth}{!}{%
\begin{tabular}{lcccclcclccc}
\toprule
 &
   &
  \multicolumn{3}{c}{General Knowledge} &
   &
  \multicolumn{2}{c}{Mathmetical Reasoning} &
   &
  \multicolumn{2}{c}{Code Generation} &
  \multicolumn{1}{c}{} \\ \cline{3-5} \cline{7-8} \cline{10-11}
\multirow{-2}{*}{Models} &
  \multirow{-2}{*}{Position} &
  \multicolumn{1}{c}{MMLU} &
  \multicolumn{1}{c}{HellaSwag} &
  \multicolumn{1}{c}{TruthfulQA} &
   &
  \multicolumn{1}{c}{GSM8K} &
  \multicolumn{1}{c}{MATH} &
   &
  \multicolumn{1}{c}{MBPP} &
  \multicolumn{1}{c}{HumanEval} &
  \multicolumn{1}{c}{\multirow{-2}{*}{Average}} \\
  \hline
 &
  \cellcolor[HTML]{EFEFEF}/ &
  \cellcolor[HTML]{EFEFEF}\textbf{59.05} &
  \cellcolor[HTML]{EFEFEF}\textbf{65.97} &
  \cellcolor[HTML]{EFEFEF}\textbf{55.69} &
  \cellcolor[HTML]{EFEFEF} &
  \cellcolor[HTML]{EFEFEF}42.53 &
  \cellcolor[HTML]{EFEFEF}9.16 &
  \cellcolor[HTML]{EFEFEF} &
  \cellcolor[HTML]{EFEFEF}49.6 &
  \cellcolor[HTML]{EFEFEF}42.7 &
  \cellcolor[HTML]{EFEFEF}{\textbf{46.37}} \\
 &
  Attention & 53.25& 57.03 & \underline{54.59} && 39.20 & 8.14 && 46.6 & 36.0 & 42.12\\
 &
  MLP &
   54.71&
   58.00&
   {53.24}&
   &
   38.51&
   8.14 &
   &
   47.6&
   35.4&
   42.23 \\
\multirow{-4}{*}{Chat} &
  Linear &
   \cellcolor[HTML]{CBCEFB}\underline{54.95}&
   \cellcolor[HTML]{CBCEFB}\underline{58.83}&
   \cellcolor[HTML]{CBCEFB}53.73&
   &
   38.82&
   8.26&
   &
   47.6&
   37.2&
   \underline{42.77}\\
   \hline
 &
  \cellcolor[HTML]{EFEFEF}/ &
  \multicolumn{1}{c}{\cellcolor[HTML]{EFEFEF}60.77} &
  \multicolumn{1}{c}{\cellcolor[HTML]{EFEFEF}58.68} &
  \multicolumn{1}{c}{\cellcolor[HTML]{EFEFEF}44.68} &
  \cellcolor[HTML]{EFEFEF} &
  \multicolumn{1}{c}{\cellcolor[HTML]{EFEFEF}\textbf{63.38}} &
  \multicolumn{1}{c}{\cellcolor[HTML]{EFEFEF}\textbf{22.74}} &
  \cellcolor[HTML]{EFEFEF} &
  \multicolumn{1}{c}{\cellcolor[HTML]{EFEFEF}38.1} &
  \multicolumn{1}{c}{\cellcolor[HTML]{EFEFEF}23.8} &
  \multicolumn{1}{c}{\cellcolor[HTML]{EFEFEF}\textbf{44.59}} \\
 &
  Attention &
  \multicolumn{1}{c}{60.11} &
  \multicolumn{1}{c}{58.48} &
  \multicolumn{1}{c}{41.13} &
   &
  \cellcolor[HTML]{CBCEFB}{\underline{58.76}} &
  \cellcolor[HTML]{CBCEFB}{\underline{19.92}} &
   &
  \multicolumn{1}{c}{41.1} &
  \multicolumn{1}{c}{25.6} &
  \multicolumn{1}{c}{\underline{43.59}} \\
 &
  MLP &
  \multicolumn{1}{c}{58.74} &
  \multicolumn{1}{c}{58.12} &
  \multicolumn{1}{c}{43.57} &
   &
  \multicolumn{1}{c}{53.37} &
  \multicolumn{1}{c}{16.68} &
   &
  \multicolumn{1}{c}{44.9} &
  \multicolumn{1}{c}{29.3} &
  \multicolumn{1}{c}{{43.53}} \\
\multirow{-4}{*}{Math} &
  Linear &
  \multicolumn{1}{c}{57.93} &
  \multicolumn{1}{c}{58.24} &
  \multicolumn{1}{c}{39.41} &
   &
  \multicolumn{1}{c}{47.16} &
  \multicolumn{1}{c}{13.9} &
   &
  \multicolumn{1}{c}{41.1} &
  \multicolumn{1}{c}{25.6} &
  \multicolumn{1}{c}{40.48} \\
  \hline
 &
  \cellcolor[HTML]{EFEFEF}/ &
  \cellcolor[HTML]{EFEFEF}50.58 &
  \cellcolor[HTML]{EFEFEF}53.19&
  \cellcolor[HTML]{EFEFEF}45.29&
  \cellcolor[HTML]{EFEFEF} &
  \cellcolor[HTML]{EFEFEF}31.69&
  \cellcolor[HTML]{EFEFEF}4.84&
  \cellcolor[HTML]{EFEFEF} &
  \cellcolor[HTML]{EFEFEF}\textbf{50.9}&
  \cellcolor[HTML]{EFEFEF}\textbf{40.9}&
  \cellcolor[HTML]{EFEFEF}{\textbf{39.63}}\\
 &
  Attention &
  \multicolumn{1}{c}{47.88} &
  \multicolumn{1}{c}{53.51} &
  \multicolumn{1}{c}{42.96} &
   &
  \multicolumn{1}{c}{27.29} &
  \multicolumn{1}{c}{5.16} &
   &
  \cellcolor[HTML]{CBCEFB}{{41.6}} &
  \cellcolor[HTML]{CBCEFB}{\underline{33.5}} &
  \multicolumn{1}{c}{{35.99}} \\
 &
  MLP &
  \multicolumn{1}{c}{45.65} &
  \multicolumn{1}{c}{54.56} &
  \multicolumn{1}{c}{42.59} &
   &
  \multicolumn{1}{c}{34.34} &
  \multicolumn{1}{c}{6.08} &
   &
  \multicolumn{1}{c}{\underline{47.9}} &
  \multicolumn{1}{c}{28.0} &
  \multicolumn{1}{c}{\underline{37.02}} \\
\multirow{-4}{*}{Code} &
  Linear &
   39.99&
   53.80&
   36.47&
   &
   22.52&
   4.82 &
   &
   40.6 &
   31.1 & 32.76\\
   \bottomrule
\end{tabular}%
}
\vspace{-0.1cm}
\caption{Impact of quantizing different module types on Mistral 7B. Compression of attention modules yields the best results for Math and Code models, while Linear module compression performs optimally for the Chat model.}
\label{tab:position_mistral}
\vspace{-0.2cm}
\end{table*}

\vspace{-0.2cm}
\paragraph{Models \& Dataets} 
Our experimental evaluation focuses on three model families: LLaMA-2 7B \cite{touvron2023llama}, Mistral 7B \cite{jiang2023mistral}, and LLaMA-2 13B \cite{touvron2023llama}, each covering distinct specializations in: general knowledge (Chat), mathematical reasoning (Math), and code generation (Code). Additional details regarding the model variants and their sources are provided in Appendix \ref{app:model_source}.

We assess performance using seven benchmark datasets across three domains: MMLU \cite{hendrycks2020measuring}, HellaSwag \cite{zellers2019hellaswag} and TruthfulQA \cite{lin2022truthfulqa} for assessing general knowledge and reasoning capabilities; GSM8K \cite{cobbe2021training} and MATH \cite{hendrycks2021measuring} for testing mathematical reasoning proficiency; HumanEval \cite{chen2021evaluating} and MBPP \cite{austin2021program} for evaluating code generation ability. To ensure consistent and unbiased assessment, model performance is evaluated using zero-shot accuracy, with pass@1 rate specifically measuring code generation correctness.

\begin{table*}[htbp]
\centering
\resizebox{\textwidth}{!}{
\begin{tabular}{lcccccccc}
\toprule
Model & MMLU & HellaSwag & TruthfulQA & GSM8K & MATH & MBPP & HumanEval & Average \\
\midrule
\rowcolor{gray!20}
Chat (1bit)           & 46.40 & \textbf{58.23} & 43.70 & \textbf{22.74} & \textbf{5.08} & 0.0 & 0.0 & \textbf{25.55} \\
Chat (SVD, $r=128$)   & \textbf{46.52} & 58.04 & \textbf{45.17} & 21.99 & 4.94 & 0.0 & 0.0 & 25.14 \\
\midrule
\rowcolor{gray!20}
Math (1bit)           & \textbf{41.03} & 56.68 & \textbf{34.52} & \textbf{47.46} & \textbf{8.42} & \textbf{23.1} & \textbf{14.0} & \textbf{32.17} \\
Math (SVD, $r=128$)   & 40.76 & \textbf{57.60} & 33.29 & 45.49 & 7.62 & 22.8 & \textbf{14.0} & 31.65 \\
\midrule
\rowcolor{gray!20}
Code (1bit)           & \textbf{41.36} & \textbf{57.81} & \textbf{32.80} & \textbf{8.19} & \textbf{3.36} & \textbf{27.3} & \textbf{11.0} & \textbf{25.97} \\
Code (SVD, $r=128$)   & 33.17 & 57.23 & 31.65 & 7.58 & 3.06 & 26.6 & 4.9 & 24.94 \\
\bottomrule
\end{tabular}
}
\caption{Performance of compressed expert models (Chat, Math, Code) under 1-bit quantization and SVD compression (r=128) applied to task vectors for LLaMA2-7B models.}
\label{tab:experts_compressed}
\vspace{-1em}
\end{table*}


\subsection{Compare to Expert Models and Traditional Merging}

Using Chat, Math, and Code fine-tuned variants of the LLaMA2-7B pretrained model, we first evaluate the performance of compressed expert models against their fine-tuned counterparts to assess the effectiveness of task-vector compression. We then compare our \texttt{1bit}-Merging model with existing task-vector-based merging methods. Results across seven datasets are summarized in Table~\ref{tab:llama2-7b}.

\paragraph{Compressed Expert Models}
The compressed expert models achieve strong performance on their respective specialized tasks and exhibit varying degrees of cross-domain generalization. As shown in Table~\ref{tab:ablation}, the \texttt{1bit}-Chat model achieves an average score of 25.55, slightly outperforming its fine-tuned counterpart (25.50). The \texttt{1bit}-Math model improves from 31.52 to 32.17, while the \texttt{1bit}-Code model increases from 24.98 to 25.97. These results suggest that module-specific 1-bit quantization preserves—and in some cases even enhances—the overall model performance across various tasks. We think this is because the quantization eliminates the impact of insignificant features.

Moreover, we further observe that the compressed models maintain strong performance beyond their primary domain.
For instance, the \texttt{1bit}-Math model also improves by 1.11 points on general knowledge and 1.25 points on code generation—areas outside its main focus.
This suggests that the removal of redundant domain knowledge by our module-specific compression strategy could improve the out-of-domain capacities while preserving the performance in the original domain, without introducing additional parameters or computational overhead.

\vspace{-0.1cm}
\paragraph{Compare to Traditional Merging}
As presented in Table~\ref{tab:llama2-7b}, our \texttt{1bit}-Merging approach achieves superior performance over traditional model merging techniques along two key dimensions.
\textbf{(1) Broad and balanced performance across domains:}
While baseline methods exhibit domain-specific strengths—TIES-Merging performs well on mathematical tasks, and DARE achieves strong results in general knowledge and code generation—our approach delivers consistently competitive performance across all domains without bias toward any single task. Specifically, \texttt{1bit}-Merging outperforms TIES-Merging by 1.45 points in mathematical reasoning and DARE by 2.71 points in general knowledge. Although we observe a modest 0.75-point drop in code generation compared to DARE, this is primarily due to the relatively low performance of the fine-tuned Code model itself. In fact, our method improves upon that baseline by 8.35 points, demonstrating substantial gains in absolute terms.
\textbf{(2) Outperforms average expert performance:}
Most notably, our method achieves an average score of 36.74 across all tasks, exceeding the average performance (35.16) of individual fine-tuned expert models. This gain of 1.58 points suggests that our merging strategy effectively preserves and combines specialized capabilities from different experts, enabling overall performance that surpasses the best individual models.

\begin{table*}[ht!]
\resizebox{\textwidth}{!}{%
\begin{tabular}{lccclcclccc}
\toprule
\multirow{2}{*}{Method} &
  \multicolumn{3}{c}{General Knowledge} &
   &
  \multicolumn{2}{c}{Mathmetical Reasoning} &
   &
  \multicolumn{2}{c}{Code Generation} &
  \multirow{2}{*}{Average} \\ 
  \cline{2-4} \cline{6-7} \cline{9-10}
                & MMLU  & HellaSwag & TruthfulQA &  & GSM8K & MATH &  & MBPP & HumanEval &  \\
                \hline
                \rowcolor{gray!10} 
\multicolumn{11}{c}{\textit{Finetuned Models}} \\
Chat & 59.05 & \textbf{65.97} & \textbf{55.69} &  & 42.53 & 9.16 &  & 49.6  & 42.7 & \textbf{46.37} \\
Math & \textbf{60.77} & 58.68 & 44.68 &  & \textbf{63.38} & \textbf{22.74} &  & 38.1 & 23.8 & 44.59 \\
Code & 50.58 & 53.19 & 45.29 &  & 31.69 & 4.84 &  & \textbf{50.9} & \textbf{40.9} & 39.63 \\
\hline
\rowcolor{gray!10} 
\multicolumn{11}{c}{\textit{Compressed Expert Models}} \\
\texttt{1bit}-Chat & 54.71 & 58.00 & \textbf{53.24} &  & 38.51 & 8.14 &  & \textbf{47.6} & \textbf{35.4} & 42.23 \\
\texttt{1bit}-Math & \textbf{60.11} & \textbf{58.48} & 41.13 &  & \textbf{58.76} & \textbf{19.92} &  & 41.1 & 25.6 & \textbf{43.59} \\
\texttt{1bit}-Code & 47.88 & 53.51 & 42.96 &  & 27.29  & 5.16 &  & 41.6 & 33.5 & 35.99 \\
\hline
\rowcolor{gray!10} 
\multicolumn{11}{c}{\textit{Model Merging Methods}} \\
Task Arithmetic & 47.34 & 46.80 & 41.00 &  & 52.16 & 13.26 &  & 32.1 & 29.9 &  39.17\\
TIES-Merging    & \textbf{57.20} & 57.59 & 48.71 &  & 55.50 & 15.00 &  & 48.4 & 41.5 &  46.30 \\
DARE & 55.36 & 55.77 & 42.84 &  & 57.39 & 15.00 &  & \textbf{49.4} & 39.0 & 44.97 \\
\rowcolor{gray!20}
\texttt{1bit}-Merging    & 56.37 & \textbf{57.87} & \textbf{53.06} &  & \textbf{60.42} & \textbf{20.60} &  & 48.6 & \textbf{42.1} & \textbf{48.43} \\
\bottomrule
\end{tabular}%
}
\vspace{-0.2cm}
\caption{Performance evaluation of merged Mistral 7B Models (Chat, Math, Code) across 7 task-specific datasets}
\label{tab:mistral-7b}
\vspace{0.55cm}
\renewcommand{\arraystretch}{0.95}
\resizebox{\textwidth}{!}{%
\begin{tabular}{lccclcclccc}
\toprule
\multirow{2}{*}{Method} &
  \multicolumn{3}{c}{General Knowledge} &
   &
  \multicolumn{2}{c}{Mathmetical Reasoning} &
   &
  \multicolumn{2}{c}{Code Generation} &
  \multirow{2}{*}{Average} \\ 
  \cline{2-4} \cline{6-7} \cline{9-10}
                & MMLU  & HellaSwag & TruthfulQA &  & GSM8K & MATH &  & MBPP & HumanEval &  \\
                \hline
                \rowcolor{gray!10} 
\multicolumn{11}{c}{\textit{Finetuned Models}} \\
Chat & \textbf{53.17} & 60.73 & \textbf{40.88} &  & 32.37 & 6.70 &  & 16.5  & 7.9 & 31.18 \\
Math & 52.73 & \textbf{61.10} & 37.09 &  & \textbf{55.50} & \textbf{10.84} &  & \textbf{28.8} & \textbf{15.9} & \textbf{37.42} \\
Code & 52.65 & 60.42 & 40.64 &  & 27.29 & 5.74 &  & 21.3 & 10.4 & 31.21 \\
\hline
\rowcolor{gray!10} 
\multicolumn{11}{c}{\textit{Compressed Expert Models}} \\
\texttt{1bit}-Chat & 53.47 & 60.40 & \textbf{41.25} &  & 31.39 & 6.58 &  & 28.1 & 5.5 & 32.38 \\
\texttt{1bit}-Math & \textbf{53.52} & \textbf{61.39} & 35.86 &  & \textbf{56.71} & \textbf{10.38} &  & \textbf{31.6} & \textbf{16.5} & \textbf{37.80} \\
\texttt{1bit}-Code & 53.06 & 60.08 & 39.17 &  & 26.61 & 6.12 &  & 21.6 & 12.2 & 31.26 \\
\hline
\rowcolor{gray!10} 
\multicolumn{11}{c}{\textit{Model Merging Methods}} \\
Task Arithmetic & 52.22 & 57.52 & 41.49 &  & 49.89 & 7.32 &  & 24.1 & 9.1 & 34.95 \\
TIES-Merging    & 55.48 & 60.65 & 39.05 &  & 52.46 & 9.90 &  & 40.4 & 21.3 & 39.89 \\
DARE & \textbf{55.65} & \textbf{61.66} & 40.51 &  & 55.19 & 9.08 &  & 39.1 & 20.1 & 40.18 \\
\rowcolor{gray!20}
\texttt{1bit}-Merging    & 55.48 & 60.22 & \textbf{42.21} &  & \textbf{56.65} & \textbf{10.40} &  & \textbf{40.6} & \textbf{22.0} & \textbf{41.07} \\
\bottomrule
\end{tabular}%
}
\vspace{-0.2cm}
\caption{Performance evaluation of merged LLaMA2-13B Models (Chat, Math, Code) across 7 task-specific datasets}
\label{tab:llama2-13b}
\vspace{-0.3cm}
\end{table*}

\subsection{Compare to Dynamic Merging Method}

We conduct additional experiments with Twin-Merging~\cite{twin-merging}, which utilizes SVD compression for task vectors and, similar to our method, employs dynamic merging to select the appropriate expert model based on input characteristics.
To ensure a comprehensive and fair comparison, we set the rank $r = 128$, matching the memory storage of 1-bit quantization. This configuration allows for a direct comparison between compression techniques under equivalent storage constraints. 
We apply both methods to the Chat, Math, and Code expert models derived from LLaMA2-7B, and evaluate their performance after merging.

The performance of compressed expert models using both 1-bit quantization and SVD compression on task vectors is summarized in Table \ref{tab:experts_compressed}.
Our results show that SVD compression fails to fully capture fine-tuning information and consistently underperforms 1-bit quantization across all expert models. Notably, low-rank approximation leads to significant degradation when compressing math and code-related task vectors. 
For instance, SVD-compressed Math models exhibit a 1.97-point drop in GSM8K accuracy, while Code models show a 6.1-point reduction in HumanEval compared to their 1-bit quantized counterparts.
As a result, combining SVD-compressed vectors with dynamic routing struggles to match the effectiveness of our 1bit-merging approach, as the fundamental information loss during compression limits the potential benefits of the routing mechanism.

\texttt{1bit}-Merging outperforms Twin-Merging by 2.09 points on GSM8K (48.52 vs. 46.43), as well as the higher MATH score (9.04 vs. 8.14), indicating stronger mathematical reasoning capabilities. In code generation, the performance gap is even more pronounced: \texttt{1bit}-Merging improves upon Twin-Merging by 5.8 points in HumanEval (18.9 vs. 13.1) and 1.0 points in MBPP (30.1 vs. 29.1). While Twin-Merging shows slightly better general knowledge performance on MMLU (48.36 vs. 47.23), our method maintains competitive results overall and surpasses Twin-Merging in task-specific domains that require precise reasoning.


\begin{table*}[htbp]
\renewcommand{\arraystretch}{0.95}
\resizebox{\textwidth}{!}{%
\begin{tabular}{lccclcclccc}
\toprule
\multirow{2}{*}{Method} &
  \multicolumn{3}{c}{General Knowledge} &
   &
  \multicolumn{2}{c}{Mathmetical Reasoning} &
   &
  \multicolumn{2}{c}{Code Generation} &
  \multirow{2}{*}{Average} \\ 
  \cline{2-4} \cline{6-7} \cline{9-10}
                & MMLU  & HellaSwag & TruthfulQA &  & GSM8K & MATH &  & MBPP & HumanEval &  \\
                \hline
LLaMA2-7B & 41.50 & 49.63 & 37.45 &  & 47.34 & 6.46 &  & 13.5 & 7.3 &  29.03\\
\rowcolor{gray!10}
\quad w/ \textit{Compressed Experts} & 42.49 & 51.45 & 39.17 &  & 45.72 & 7.70 &  & 15.8 & 8.5 &  30.09 \textcolor{blue}{(+1.06)}\\
Mistral-7B & 47.34 & 46.80 & 41.00 &  & 52.16 & 13.26 &  & 32.1 & 29.9 &  39.17\\
\rowcolor{gray!10}
\quad w/ \textit{Compressed Experts} & 45.98 & 48.03 & 40.76 &  & 46.32 & 12.14 &  & 33.8 & 29.3 & 36.62 \textcolor{red}{(-2.55)}\\
LLaMA2-13B  & 52.22 & 57.52 & 41.49 &  & 49.89 & 7.32 &  & 24.1 & 9.1 & 34.95 \\
\rowcolor{gray!10}
\quad w/ \textit{Compressed Experts} & 52.57 & 58.01 & 41.49 &  & 53.60 & 7.48 &  & 27.3 & 12.2 & 36.09 \textcolor{blue}{(+1.14)} \\
\bottomrule
\end{tabular}%
}
\caption{Performance comparison between merging original and compressed fine-tuned models.}
\label{tab:bit-merge}
\vspace{-0.4cm}
\end{table*}

\vspace{-0.3em}
\subsection{Scalability to Architecture and Size}
\paragraph{Generalization to Different Architectures}
To assess the architectural generality of our method, we apply \texttt{1bit}-Merging to expert models based on Mistral 7B. As shown in Table~\ref{tab:position_mistral} and Table \ref{tab:mistral-7b}, our method consistently performs well across domains. \texttt{1bit}-Merging outperforms merging strategies—Task Arithmetic (+9.26), TIES-Merging (+2.13), and DARE (+3.46) on average across seven datasets. 
For mathematical reasoning, \texttt{1bit}-Merging substantially surpasses TIES-Merging by 4.92 points on GSM8K and 5.6 points on MATH, demonstrating its strong capability in preserving reasoning ability.


\vspace{-0.3em}
\paragraph{Scaling to Larger Model Size}
To evaluate the scalability of our method, we apply \texttt{1bit}-Merging to three fine-tuned variants of LLaMA-2 13B: Chat, Math, and Code. As shown in Table~\ref{tab:llama2-13b} and Table \ref{tab:position_llama_13}, our approach remains effective at this larger scale, outperforming all baseline methods on average—by 6.12 points over Task Arithmetic, 1.18 points over TIES-Merging, and 0.89 points over DARE.
Notably, \texttt{1bit}-Merging achieves a 14.54\% relative improvement (1.32 points) over DARE on MATH and a 4.20\% relative gain (1.70 points) on TruthfulQA. In code generation, it consistently surpasses TIES-Merging by 0.45 points across MBPP and HumanEval.


\section{Further Analyses}

\paragraph{Ablation Studies}
Merging compressed expert models is effective for Task Arithmetic, but shows limitations when applied to parameter-dropping methods such as TIES-Merging and DARE (see Table~\ref{tab:bit-merge}). This is likely due to the fact that parameter-dropping techniques disrupt the structured knowledge distributions critical for task-specific performance.
In our experiments on LLaMA2-7B and LLaMA2-13B, we observe that compressed expert models already achieve slightly better performance than their individually fine-tuned counterparts. Merging these compressed experts further improves performance by 1.06 and 1.14 points, respectively, over merging the fine-tuned models.

\begin{figure}[htbp]
    \centering
    \includegraphics[width=0.85\linewidth]{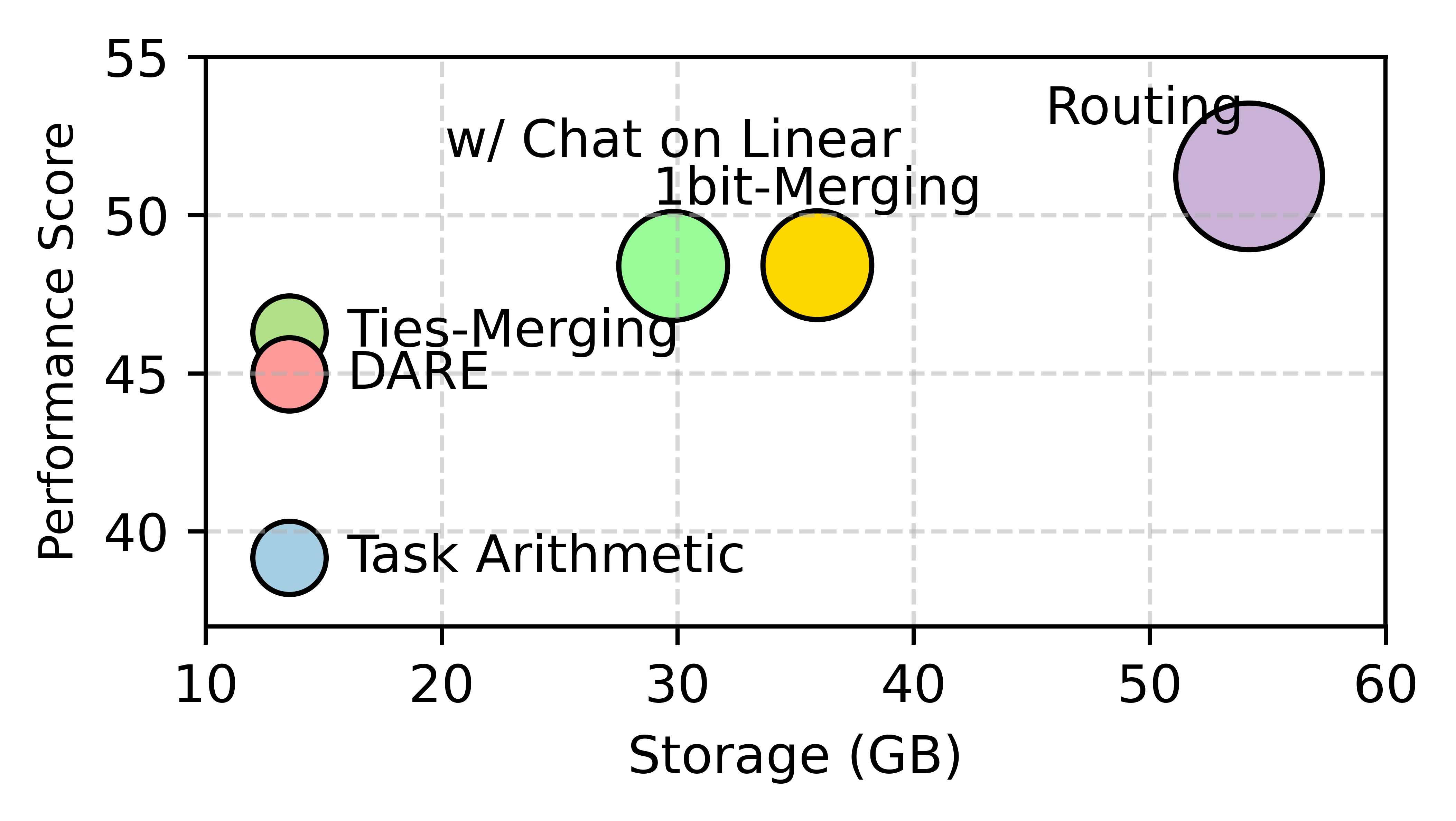}
    \vspace{-3mm}
    \caption{Performance vs. storage trade-offs for Mistral 7B deployment. Task-specific routing (Routing) achieves strong performance but requires full parameter storage. Model merging reduces storage at the cost of performance. \texttt{1bit}-Merging strikes a better balance, and applying 1-bit quantization to all linear layers in Chat (``w/ Chat on Linear'') further improves efficiency.}
    \label{fig:storage}
    \vspace{-5mm}
\end{figure}

\vspace{-0.3em}
\paragraph{Performance vs. Storage Trade-offs}
Figure~\ref{fig:storage} illustrates the trade-off between performance and storage requirements among different model deployment strategies, using Mistral 7B expert models. Task-specific routing (Routing), which maintains full parameters for each expert, achieves performance comparable to individually fine-tuned models and serves as an upper bound. However, this comes at a significantly increased storage cost.
Model merging methods reduce storage demands, but generally underperform Routing by more than 4.93 points on average. In contrast, our \texttt{1bit}-Merging strikes a favorable balance—retaining 94.53\% of the performance of Routing while requiring only 66.25\% of its storage. Notably, we observe that Chat models are less sensitive to the choice of quantized modules. Leveraging this, we apply 1-bit quantization to all linear layers in the Chat model, further reducing storage usage to 55.02\% with minimal impact on performance.


\section{Conclusion}

We propose \texttt{1bit}-Merging, a novel framework that effectively combines specialized language models while addressing the fundamental trade-off between performance and storage efficiency. By incorporating dynamic routing with binary quantization, our approach maintains task-specific expertise while significantly reducing storage overhead. Extensive experiments across general knowledge, mathematical reasoning, and code generation tasks demonstrate that \texttt{1bit}-Merging not only preserves the specialized capabilities of individual models but often enhances their performance. 

\newpage

\section*{Limitations}
Our approach faces limitations that are both shared with existing merging methods. (1) At a theoretical level, despite the empirical success of weight interpolation techniques, we still lack a comprehensive understanding of why and when these methods work effectively. Although recent research has revealed interesting patterns about weight disentanglement and cross-task linear relationships, the precise conditions for optimal weight interpolation remain unclear.
(2) A practical constraint lies in the architectural requirements of our method. Currently, the merging process is restricted to models sharing identical architectures, as it relies on direct parameter operations between models. This limitation becomes particularly challenging when seeking to merge models with specific capabilities, as finding suitable fine-tuned models with matching architectures can be difficult. Looking ahead, extending model merging capabilities to heterogeneous architectures represents an exciting and challenging frontier for future research, though this would require developing novel techniques for mapping and aligning parameters across different architectural designs.

\section*{Ethics Statement}
This study utilizes publicly available datasets for our models. Prior research endeavors have generally taken ethical considerations into account. We have manually inspected a subset of samples and found no explicit ethical concerns, including violent or offensive content. Nonetheless, it is crucial to highlight that the output generated by large language models lacks the degree of control we might assume. Consequently, we are prepared to implement measures to mitigate any unforeseen outputs.

\bibliography{custom}

\appendix

\section{Appendix}
\label{sec:appendix}

\subsection{Implementation Details}
\label{appen:implement_detail}

\paragraph{Router} We implement the router as a three-layer linear network with Leaky ReLU activations and batch normalization, serving as an input classifier that categorizes each input instance into one of the three task types: math, code, or chat.
To train the router, we construct a dedicated validation dataset by randomly sampling instances from the respective downstream datasets and combining them. This ensures that the training data is representative of the diverse input distributions the router is expected to encounter during inference.
For the training process, we used a learning rate of 5e-5 and trained the model for 20 epochs on the validation set. The validation set was taken from a split of the overall training set, and we used 800 samples for each downstream dataset during the router training.

\paragraph{Hyperparameters} 
For Task Arithmetic, we use a default scaling coefficient of $\lambda = 1$, which maintains the original magnitude of task vector when adding the pretrained backbone. 
However, the DARE method has been observed to be more sensitive to variations in both the scaling coefficient \(\lambda\) and the drop rate parameter \(r\). 
To achieve a balanced performance, we set the scaling coefficient to \(\lambda = 0.5\) and establish a default drop rate of \(r = 0.5\) for DARE.
Similarly, for TIES-Merging, which requires the specification of a masking ratio, we set the default mask ratio to \(r = 0.7\) across all experiments. 

\subsection{Model Variants and Sources}
\label{app:model_source}
We evaluate our method across multiple fine-tuned variants of LLaMA2-7B, Mistral 7B, and LLaMA2-13B models, focusing on specialized capabilities in instruction-following (Chat), mathematical reasoning (Math), and code generation (Code). Table~\ref{tab:model_checkpoints_l2_7b} summarizes the fine-tuned variants used for LLaMA2-7B along with their corresponding Hugging Face identifiers and use cases. For completeness, we also include similarly fine-tuned versions from other base models: Mistral 7B (Table~\ref{tab:model_checkpoints_mis_7b}) and LLaMA2-13B Table~\ref{tab:model_checkpoints_l2_13b}. These variants are obtained from publicly available checkpoints and are used consistently across all compression and merging experiments reported in this work.

\begin{table*}[ht!]
\centering
\small
\begin{tabular}{lcccc}
\toprule
\textbf{} & \textbf{Model Type} & \textbf{Hugging Face Identifier} & \textbf{Use Case} \\
\midrule
\multirow{4}{*}{LLaMA2-7B} & Pretrained Base Model & meta-llama/Llama-2-7b-hf & General language modeling \\
& Chat-Tuned Model        & meta-llama/Llama-2-7b-chat-hf & Dialogue and instruction-following \\
& Math-Tuned Model        & TIGER-Lab/MAmmoTH-7B & Mathematical reasoning \\
& Code-Tuned Model        & mrm8488/llama-2-coder-7b & Code generation \\
\bottomrule
\end{tabular}
\caption{Fine-tuned variants and corresponding Hugging Face identifiers for LLaMA-2 7B models.}
\label{tab:model_checkpoints_l2_7b}
\end{table*}

\begin{table*}[htbp]
\centering
\small
\begin{tabular}{lcccc}
\toprule
\textbf{} & \textbf{Model Type} & \textbf{Hugging Face Identifier} & \textbf{Use Case} \\
\midrule
\multirow{4}{*}{Mistral 7B} & Pretrained Base Model & mistralai/Mistral-7B-v0.1 & General language modeling \\
& Chat-Tuned Model        & mistralai/Mistral-7B-Instruct-v0.1 & Dialogue and instruction-following \\
& Math-Tuned Model        & TIGER-Lab/MAmmoTH2-7B & Mathematical reasoning \\
& Code-Tuned Model        & Nondzu/Mistral-7B-codealpaca-lora & Code generation \\
\bottomrule
\end{tabular}
\caption{Fine-tuned variants and corresponding Hugging Face identifiers for Mistral 7B models.}
\label{tab:model_checkpoints_mis_7b}
\end{table*}


\begin{table*}[htbp]
\centering
\small
\begin{tabular}{lcccc}
\toprule
\textbf{} & \textbf{Model Type} & \textbf{Hugging Face Identifier} & \textbf{Use Case} \\
\midrule
\multirow{4}{*}{LLaMA2-13B} & Pretrained Base Model & meta-llama/Llama-2-13b-hf & General language modeling \\
& Chat-Tuned Model        & meta-llama/Llama-2-13b-hf & Dialogue and instruction-following \\
& Math-Tuned Model        & TIGER-Lab/MAmmoTH-13B & Mathematical reasoning \\
& Code-Tuned Model        & emre/llama-2-13b-code-chat & Code generation \\
\bottomrule
\end{tabular}
\caption{Fine-tuned variants and corresponding Hugging Face identifiers for LLaMA-2 13B models.}
\label{tab:model_checkpoints_l2_13b}
\end{table*}

\subsection{Additional Experiment Details}

\paragraph{Layer Types on Compression Performance}
We further analyze how the choice of compression module impacts performance on LLaMA-2 13B architecture. As shown in Table~\ref{tab:position_llama_13}, Chat models exhibit minimal variation in performance across different compression strategies—LLaMA-2 13B favors the attention layers. In contrast, Math and Code models benefit consistently from preserving attention layers, with quantization of these components yielding the smallest performance degradation. This suggests that domain-specific knowledge for complex reasoning tasks is more concentrated in the attention modules, making them less tolerant to aggressive compression.

\begin{table*}[!h]
\renewcommand{\arraystretch}{0.95}
\resizebox{\textwidth}{!}{%
\begin{tabular}{lcccclcclccc}
\toprule
 &
   &
  \multicolumn{3}{c}{General Knowledge} &
   &
  \multicolumn{2}{c}{Mathmetical Reasoning} &
   &
  \multicolumn{2}{c}{Code Generation} &
  \multicolumn{1}{c}{} \\ \cline{3-5} \cline{7-8} \cline{10-11}
\multirow{-2}{*}{Models} &
  \multirow{-2}{*}{Position} &
  \multicolumn{1}{c}{MMLU} &
  \multicolumn{1}{c}{HellaSwag} &
  \multicolumn{1}{c}{TruthfulQA} &
   &
  \multicolumn{1}{c}{GSM8K} &
  \multicolumn{1}{c}{MATH} &
   &
  \multicolumn{1}{c}{MBPP} &
  \multicolumn{1}{c}{HumanEval} &
  \multicolumn{1}{c}{\multirow{-2}{*}{Average}} \\
  \hline
 &
  \cellcolor[HTML]{EFEFEF}/ &
  \cellcolor[HTML]{EFEFEF}{53.17} &
  \cellcolor[HTML]{EFEFEF}{60.73} &
  \cellcolor[HTML]{EFEFEF}\underline{40.88} &
  \cellcolor[HTML]{EFEFEF} &
  \cellcolor[HTML]{EFEFEF}32.37 &
  \cellcolor[HTML]{EFEFEF}6.70 &
  \cellcolor[HTML]{EFEFEF} &
  \cellcolor[HTML]{EFEFEF}16.5 &
  \cellcolor[HTML]{EFEFEF}7.9 &
  \cellcolor[HTML]{EFEFEF}{31.18} \\
 &
  Attention & \cellcolor[HTML]{CBCEFB}\textbf{53.47}& \cellcolor[HTML]{CBCEFB}60.40 & \cellcolor[HTML]{CBCEFB}\textbf{41.25} && 31.39 & 6.58 && 28.1 & 5.5 & \underline{32.38}\\
 &
  MLP &
   53.13&
   60.94&
   38.68&
   &
   31.39&
   6.58 &
   &
   20.1&
   9.1&
   31.42 \\
\multirow{-4}{*}{Chat} &
  Linear &
   \underline{53.38}&
   \textbf{61.30}&
   39.05&
   &
   31.08&
   6.20&
   &
   32.3&
   4.3&
   \textbf{32.52}\\
   \hline
 &
  \cellcolor[HTML]{EFEFEF}/ &
  \multicolumn{1}{c}{\cellcolor[HTML]{EFEFEF}52.73} &
  \multicolumn{1}{c}{\cellcolor[HTML]{EFEFEF}61.10} &
  \multicolumn{1}{c}{\cellcolor[HTML]{EFEFEF}37.09} &
  \cellcolor[HTML]{EFEFEF} &
  \multicolumn{1}{c}{\cellcolor[HTML]{EFEFEF}\underline{55.50}} &
  \multicolumn{1}{c}{\cellcolor[HTML]{EFEFEF}\textbf{10.84}} &
  \cellcolor[HTML]{EFEFEF} &
  \multicolumn{1}{c}{\cellcolor[HTML]{EFEFEF}28.8} &
  \multicolumn{1}{c}{\cellcolor[HTML]{EFEFEF}15.9} &
  \multicolumn{1}{c}{\cellcolor[HTML]{EFEFEF}{37.42}} \\
 &
  Attention &
  \multicolumn{1}{c}{53.52} &
  \multicolumn{1}{c}{61.39} &
  \multicolumn{1}{c}{35.86} &
   &
  \cellcolor[HTML]{CBCEFB}{\textbf{56.71}} &
  \cellcolor[HTML]{CBCEFB}{\underline{10.38}} &
   &
  \multicolumn{1}{c}{31.6} &
  \multicolumn{1}{c}{16.5} &
  \multicolumn{1}{c}{\textbf{37.80}} \\
 &
  MLP &
  \multicolumn{1}{c}{54.19} &
  \multicolumn{1}{c}{61.70} &
  \multicolumn{1}{c}{35.25} &
   &
  \multicolumn{1}{c}{53.98} &
  \multicolumn{1}{c}{9.10} &
   &
  \multicolumn{1}{c}{33.6} &
  \multicolumn{1}{c}{13.4} &
  \multicolumn{1}{c}{{37.32}} \\
\multirow{-4}{*}{Math} &
  Linear &
  \multicolumn{1}{c}{54.91} &
  \multicolumn{1}{c}{61.89} &
  \multicolumn{1}{c}{35.62} &
   &
  \multicolumn{1}{c}{53.90} &
  \multicolumn{1}{c}{8.80} &
   &
  \multicolumn{1}{c}{32.8} &
  \multicolumn{1}{c}{14.6} &
  \multicolumn{1}{c}{\underline{37.50}} \\
  \hline
 &
  \cellcolor[HTML]{EFEFEF}/ &
  \cellcolor[HTML]{EFEFEF}52.65 &
  \cellcolor[HTML]{EFEFEF}60.42&
  \cellcolor[HTML]{EFEFEF}40.64&
  \cellcolor[HTML]{EFEFEF} &
  \cellcolor[HTML]{EFEFEF}27.29&
  \cellcolor[HTML]{EFEFEF}5.74&
  \cellcolor[HTML]{EFEFEF} &
  \cellcolor[HTML]{EFEFEF}\underline{21.3}&
  \cellcolor[HTML]{EFEFEF}10.4&
  \cellcolor[HTML]{EFEFEF}31.21\\
 &
  Attention &
  \multicolumn{1}{c}{53.06} &
  \multicolumn{1}{c}{60.08} &
  \multicolumn{1}{c}{39.17} &
   &
  \multicolumn{1}{c}{26.61} &
  \multicolumn{1}{c}{6.12} &
   &
  \cellcolor[HTML]{CBCEFB}\textbf{21.6} &
  \cellcolor[HTML]{CBCEFB}{\textbf{12.2}} &
  \multicolumn{1}{c}{\textbf{31.26}} \\
 &
  MLP &
  \multicolumn{1}{c}{53.06} &
  \multicolumn{1}{c}{60.34} &
  \multicolumn{1}{c}{38.19} &
   &
  \multicolumn{1}{c}{22.29} &
  \multicolumn{1}{c}{4.76} &
   &
  \multicolumn{1}{c}{5.5} &
  \multicolumn{1}{c}{\underline{12.2}} &
  \multicolumn{1}{c}{{28.05}} \\
\multirow{-4}{*}{Code} &
  Linear &
   53.42&
   60.60&
   37.09&
   &
   15.84&
   4.58 &
   &
   7.8 &
   8.5 & 26.83 \\
   \bottomrule
\end{tabular}%
}
\vspace{-0.3cm}
\caption{Impact of quantizing different module types on LLaMA2-13B. Compression of attention modules yields the best results for Chat, Math and Code models.}
\label{tab:position_llama_13}
\vspace{-0.3cm}
\end{table*}

\end{document}